\documentclass[letterpaper]{article}
\usepackage{aaai}
\usepackage{times}
\usepackage{helvet}
\usepackage{courier}
\usepackage{graphicx}
\usepackage{bm}
\usepackage{amssymb}
\usepackage{amsmath}
\nocopyright 
 
\begin{document}
\setcounter{secnumdepth}{2}
\title{An Introductory Survey on Attention Mechanisms in NLP Problems}
\author{Dichao Hu\\
College of Computing, Georgia Institute of Technology\\
801 Atlantic Dr NW\\
Atlanta, Georgia 30332\\
}
\maketitle
\begin{abstract}
First derived from human intuition, later adapted to machine translation for automatic token alignment, attention mechanism, a simple method that can be used for encoding sequence data based on the importance score each element is assigned, has been widely applied to and attained significant improvement in various tasks in natural language processing, including sentiment classification, text summarization, question answering, dependency parsing, etc. In this paper, we survey through recent works and conduct an introductory summary of the attention mechanism in different NLP problems, aiming to provide our readers with basic knowledge on this widely used method, discuss its different variants for different tasks, explore its association with other techniques in machine learning, and examine methods for evaluating its performance.
\end{abstract}
\section{Introduction}
\begin{figure}
\centering
\includegraphics*[width = 0.4\textwidth]{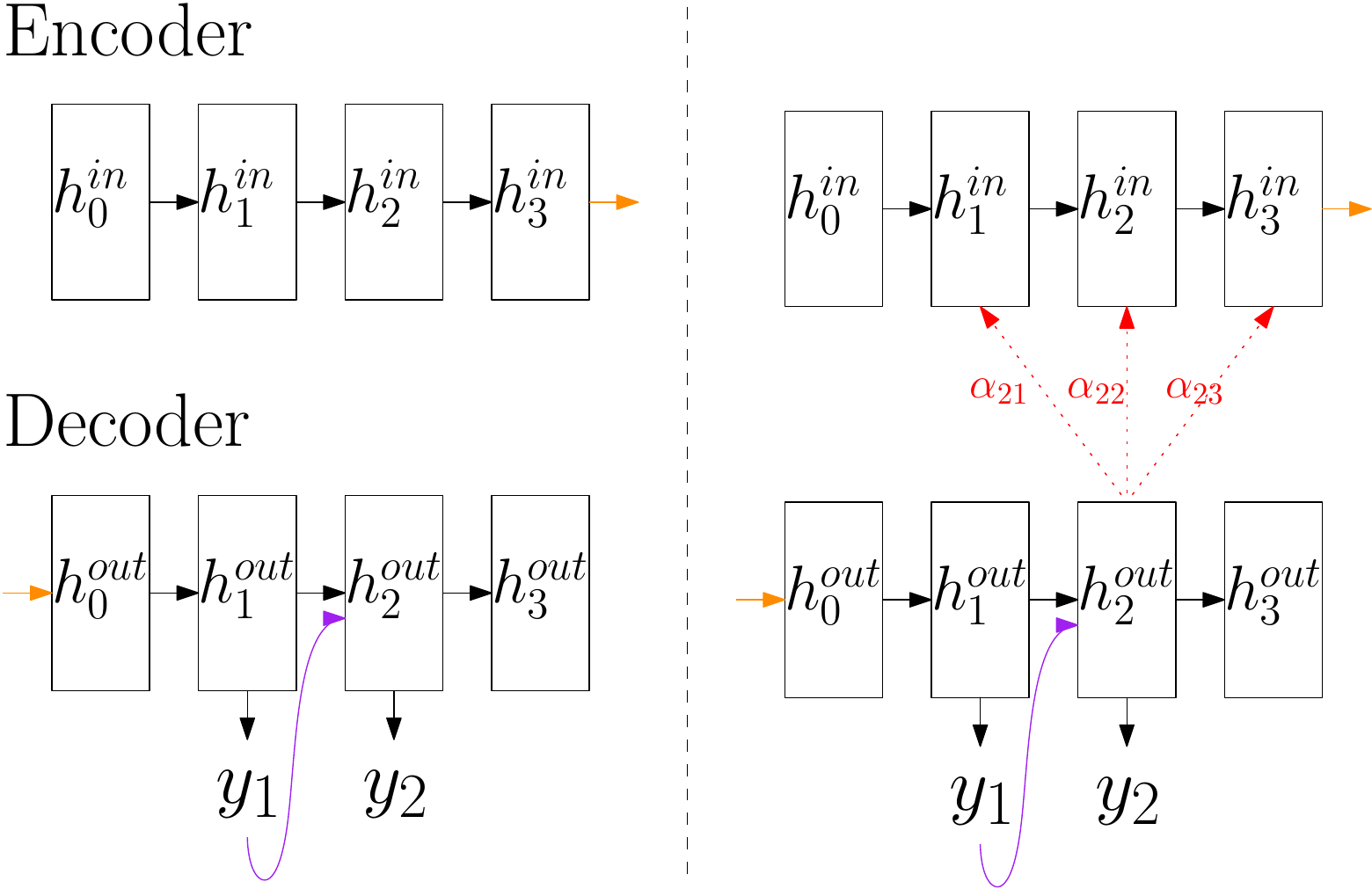}
\caption{A comparison between the traditional encoder-decoder architecture (left) and the attention-based architecture (right). During decoding, an additional attention score $\alpha_{ji}$ is computed for each source token $\bm{h_i^{in}}$ with respect to the target token $\bm{h_j^{out}}$, and the scores are then used to obtain the contextual encoding $\bm{c_j}$.}
\label{fig:comparison}
\end{figure}

We introduce our main topic via a concrete example of neural machine translation. Traditional methods \cite{kalchbrenner2013recurrent,cho2014properties} are formulated by an encoder-decoder architecture, both of which are recurrent neural networks. An input sequence of source tokens is first fed into the encoder, of which the last hidden representation is extracted and used to initialize the hidden state of the decoder, and then target tokens are generated one after another. Despite achieving higher performance compared to purely statistical methods, the RNN-based architecture suffers from two serious drawbacks. First, RNN is forgetful, meaning that old information gets washed out after being propagated over multiple time steps. Second, there is no explicit word alignment during decoding and therefore focus is scattered across the entire sequence. Aiming to resolve the issues above, attention mechanism was first introduced into neural machine translation \cite{bahdanau2014neural}. They maintain the same RNN encoder, for each step $j$  during decoding they compute an attention score $\alpha_{ji}$ for hidden representation $\bm{h_i^{in}}$ of each input token to obtain a context vector $\bm{c_j}$ (see Figure \ref{fig:comparison}):
\begin{eqnarray}\label{attention_nmt}
e_{ji} = a(\bm{h_i^{in}},\bm{h_j^{out}})\\
\alpha_{ji} = \frac{e_{ji}}{\sum_{i}{e_{ji}}}\\
\bm{c_j} = \sum_{i}{\alpha_{ji}\bm{h_i^{in}}}
\end{eqnarray} Here $\bm{c_j}$, a weighted average of elements in the input sequence, is the encoded sentence representation with respect to the current element $\bm{h_j^{out}}$, and
$a(\bm{h_i^{in}},\bm{h_j^{out}})$ is the alignment function that measures similarity between two tokens and will be discussed in detail later. Then $\bm{c_j}$ is combined with the current hidden state $\bm{h_j}$ and the last target token $\bm{y_{j-1}}$ to generate the current token $\bm{y_j}$:
\begin{eqnarray}
\bm{y_j}=f_y(\bm{h_j^{out}},\bm{y_{j-1}},\bm{c_j})\\
\bm{h_{j+1}^{out}}= f_{h}(\bm{h_j^{out}},\bm{y_{j}})
\end{eqnarray} Here $f_y$ and $f_h$ stands for the output layer and hidden layer in recurrent networks. This procedure is repeated for each token $\bm{y_j}$ until the end of the output sequence.
By introducing this additional encoding step, problems mentioned earlier can be tackled: the bad memory of RNN is no longer an issue, since the computation of attention score is performed on each element in the input sequence and therefore computation of the encoded representation $\bm{c_j}$ is unaffected by the sequence length; on the other hand, soft alignment across the input sequence can be achieved since each element is either highlighted or down-weighted based on its attention score and focus is paid only to the important parts in the sequence, discarding useless or irrelevant parts. As a result, the attention-based translation model achieved great success in machine translation and then attention mechanism got widely applied to other NLP tasks. Moreover, different variants of the basic form of attention have been proposed to handle more complex tasks. As an overview of the following sections, we will: \begin{enumerate}
\item Explain the basic form of attention in detail. (Formulation)
\item Discuss different variants based on the  special task they are dealing with. (Variation)
\item Explore how attention is associated with other concepts or techniques in machine learning, such as pre-training and ensemble. (Application)
\item Examine methods for evaluating the performance of attention. (Evaluation)
\end{enumerate}

\section{Formulation}
Previously, we discussed the application of attention in neural machine translation. To formally generalize it into the basic form of attention, we define $V$ = \{$\bm{v_i}$\} $\in \mathbb{R}^{n \times d_v}$ as a sequence of vector elements and rewrite the previous steps as:\begin{eqnarray*}
e_i=a(\bm{u,v_i}) & & (\textnormal{compute attention scores})\\
\alpha_{i} = \frac{e_{i}}{\sum_{i}{e_{i}}} & & (\textnormal{normalize})\\
c = \sum_{i}{\alpha_{i}\bm{v_i}} & & (\textnormal{encode})
\end{eqnarray*}   In the first step, $\bm{u} \in \mathbb{R}^{d_u}$ is a task-specific pattern vector to match with each element in the sequence \{$\bm{v_i}$\} based on the alignment function $a\bm{(u,v)}$ that outputs a scalar score $e_i \in \mathbb{R}$ to indicate quality of match. In most cases we assume $d_u = d_v=d$.
Common choices \cite{luong2015effective} are given by: \\
Multiplicative:
\begin{eqnarray}
\bm{u^Tv} & & \textnormal{(dot)}\\
\bm{u^T}W\bm{v}  & &  \textnormal{(general)}
\end{eqnarray}
Additive:  \begin{eqnarray}
\bm{w_2^T}tanh(W_1[\bm{u};\bm{v}])
\end{eqnarray}
MLP: \begin{eqnarray}
\sigma (\bm{w_2^T}tanh(W_1[\bm{u};\bm{v}]+\bm{b_1})+b_2)
\end{eqnarray} all of which measures pairwise similarity in different representations. The final score $\alpha_i \in \mathbb{R}$ is the normalized weight for each element $\bm{v_i}$ and then used to encode the entire sequence into a context vector $\bm{c} \in \mathbb{R}^{d_u}$, which is later incorporated into a downstream task as an additional contextual feature. Intuitively, $\bm{v_i}$ that closely matches the pattern $\bm{u}$ receives a large weight and therefore dominates the final encoding $\bm{c}$. In machine translation, the attention score can be naturally interpreted as the alignment measure of a target token to each source token. Ideally, when generating a target token $\bm{y_j}$, we expect the gold alignment source token(s) to receive high attention score and therefore the encoded representation $\bm{c_j}$ can provide closely relevant information for decoding.\\To avoid confusion, we explain our use of notation and abbreviation before our further discussion. We use lower-case symbols for scalars, lower-case boldface symbols for vectors, upper-case symbols for matrices, and upper-case boldface symbols for tensors. When we refer to attention scores, $e$ stands for original scores and $\alpha$ stands for normalized scores. $W, b$ are by default weights and biases to be learned during training, and $\bm{v_i}$ is by default a general element that can refer to either one of the following depend on the scenario: an embedded token within a sentence, a hidden representation of the token, an embedded sentence within a document, etc. For simplicity, we occasionally use $softmax$ to represent the normalization step that involves exponentiation, summation and division.

\section{Variation}
Previously, we discussed basic form of the attention mechanism. Because of its simplicity and interpretability, it is widely used in various NLP tasks. Nevertheless, such attention mechanism is in general not powerful enough for more complicated tasks. Here is a simple example \cite{sukhbaatar2015end}:\\\\ 
Sam walks into the kitchen. (1)\\
Sam picks up an apple. (2)\\
Sam walks into the bedroom. (3)\\
Sam drops the apple. (4)\\
Q: Where is the apple?\\
A: Bedroom\\\\
The underlying difficulty is that there is no direct relationship between Q and (3), and therefore we need to design a more sophisticated mechanism to guide attention to the correct location using latent clues within the context (temporal reasoning in this case). As different variants of attention have been proposed in recent years, we summarize them into several most representative categories (see Table \ref{type_of_attention}): basic attention, multi-dimensional attention, hierarchical attention, self-attention, memory-based attention and task-specific attention. From left to right, the corresponding task increases in complexity, and the mechanism becomes more task specific. We will discuss each category in detail in the following sections.\\
\begin{table}
\centering
\resizebox{0.4\textwidth}{!}{
\begin{tabular}{c|c}
Type of Attention & Purpose in brief \\ \hline
Basic Attention &Extracting important elements from a sequence\\ \hline
Multi-dimensional Attention &Capturing multiple types of interaction between terms \\ \hline
Hierarchical Attention & Extracting globally and locally important information\\ \hline
Self-Attention & Capturing deep contextual information within a sentence \\ \hline
Memory-based Attention & Discovering latent dependencies in sophisticated NLP tasks 
\\ \hline
Task-specific Attention & Capturing important information specified by the task\\ \hline

\end{tabular}}
\caption{An overview of each type of attention and its corresponding purpose} 
\label{type_of_attention}
\end{table} 
\subsection{Multi-dimensional Attention}
The basic form of attention mechanism computes a scalar score $\alpha_i$ for each term in a sequence $V$ = \{$\bm{v_i}$\}. This could be named as 1D attention or vector attention because the concatenated output scores $\bm{\alpha}$ = \{$\alpha_i$\} is a vector $\bm{\alpha} \in R^n$. The motivation for multi-dimensional attention is simply to capture multiple interaction between terms in different representation space, which can be easily constructed by directly stacking together multiple single dimensional representations. 
\begin{table}
\centering
\resizebox{0.4\textwidth}{!}{
\begin{tabular}{c|c}
1D Attention & 2D Attention \\ \hline
$\bm{u^T}W\bm{v}$ $(W \in \mathbb{R}^{d\times d})$ & $\bm{u^TWv}$ $(\bm{W} \in \mathbb{R}^{k\times d\times d})$\\ \hline
$\bm{w_2^T} tanh(W_1[\bm{u};\bm{v}])$ $(\bm{w_2} \in \mathbb{R}^e)$ & $W_2^Ttanh(W_1[\bm{u};\bm{v}])$ $(W_2 \in \mathbb{R}^{e\times k})$\\ \hline
$\sigma (\bm{w_2^T}tanh(W_1[\bm{u};\bm{v}]+\bm{b_1})+b_2)$ &
 $\sigma (W_2^Ttanh(W_1[\bm{u};\bm{v}]+\bm{b_1})+\bm{b_2})$
\\ \hline

\end{tabular}}
\caption{Some examples of extending from 1D attention to 2D, assuming $\bm{u,v} \in \mathbb{R}^d, W_1 \in \mathbb{R}^{e \times 2d}$} 
\label{2d_attention}
\end{table}\\
As an example of 2D attention in aspect and opinion terms extraction \cite{wang2017coupled}, given a sequence of hidden representation of tokens $V = \{\bm{v_i}\}$, an aspect prototype vector $\bm{u}$, and a 3D tensor $\bm{W}=\{W_k\}   \in \mathbb{R}^{K \times d \times d}$ where each slice $W_k \in \mathbb{R}^{d \times d}$ is a matrix that captures one type of composition between each token and the prototype, the vector attention score $\bm{e_i}$ (not normalized) for each element $\bm{v_i}$ is given by:
\begin{equation}
\bm{e_i} = tanh(\bm{u^TWv_i}) = concat(tanh(\bm{u^T}W_k\bm{v_i}))
\end{equation}
Then the 2D attention representation $E$ for the entire sequence is obtained by concatenation: $E=\{\bm{e_i}\}$.  As a concrete example, consider the sentence "\textit{Fish burger is the best dish}", here \textit{Fish burger} is an aspect term and therefore will receive high attention score with respect to the aspect prototype $\bm{u}$, and the learned $\bm{W}$ should be able to highlight both \textit{Fish} and \textit{burger} after they are projected to different spaces. 
A drawback of this multi-dimension approach is that a strongly indicative element can capture multiple types of attention at the same time and therefore reduces its representation power. To compensate for this, we can impose a regularization penalty in Frobenius norm on the 2D attention matrix $A$ \cite{du2018multi,lin2017structured}:
\begin{equation}
||AA^T-I||^2_F
\end{equation}
to constrain each attention column to focus on different parts in the sequence. 
\subsection{Hierarchical Attention}
Let us begin our discussion on hierarchical attention by looking at an example in document classification. Given a short document \cite{yang2016hierarchical}: 
\textit{\underline{How do I get rid of all the old} \textbf{web searches} I have on my \textbf{web browser}?
I want to clean up my \textbf{web browser} go to tools $\rightarrow$ options
Then click "delete history" and "clean up temporary internet files."}, the task is to classify the document into one of several categories, which in this case is \textit{Computer and Internet}.
Intuitively, we can identify words that provide clues for classification, since the term \textit{\textbf{web browser}} can appear frequently in computer-related documents.
Moreover, typical sentences can be potentially informative for classification as well, as non-professionals using a computer software for the first time tend to seek instruction on \textit{how to get rid of \dots} If we think of the nested structure of textual data: $character \in word \in sentence \in document$, a hierarchical attention can be constructed accordingly, either bottom-up (i.e, word-level to sentence-level) or top-down (word-level to character-level) to identify clues or extract important information both globally and locally.\\
Bottom-up construction has been used in document classification \cite{yang2016hierarchical}. Two BiGRUs are applied to generate a word-level and a sentence-level contextual representation, respectively. Then a pair of hierarchical attention layers are applied to obtain a word-level and sentence-level encoding:
\begin{eqnarray}
\bm{h_i^{(t)}}=BiGRU(\bm{v_i^{(t)}})\\
\bm{v_{i}} = \sum_{t}{softmax(\bm{u_w^T} \bm{h_i^{(t)}})\cdot \bm{h_i^{(t)}}}\\
\bm{h_{i}} = BiGRU(\bm{v_{i}})\\
\bm{c} = \sum_{i}{softmax(\bm{u_s^T} \bm{h_{i}})\cdot \bm{h_{i}}}
\end{eqnarray}
$softmax$ is equivalent to the normlization step we previously discussed. $\bm{h_i^{(t)}}$ and $\bm{h_{i}}$ stand for hidden representation for words and sentences. $\bm{u_w^T}$ and $\bm{u_s^T}$ are word-level and sentence-level pattern vectors to be learned during training. The final sentence-level representation $\bm{c}$ is then fed into a logistic regression layer to predict the category.\\   
Another type of hierarchical attention takes a top-down approach, an example of which is for grammatical error correction \cite{ji2017nested}. Consider a corrupted sentence: \textit{I have no enough previleges}. The idea is to first bulid an encoder-decoder architecture similar to the one for machine translation, then apply a word-level attention for global grammar and fluency error correction (\textit{I have no enough} $\rightarrow$ \textit{I don't have enough}), and optionally a character-level attention for local spelling error correction \textit{(previleges $\rightarrow$ privileges)}. Top-down techniques are also used in album summarization \cite{yu2017hierarchically}, where a photo-level attention is used to select appropriate photos from an album and a word-level attention is integrated with a sequence model for text generation.	
\subsection{Self Attention}\label{self-attention}
Let us revisit the steps for constructing the basic form of attention. Given a sequence of elements $V$ = \{$\bm{v_i}$\} and a pattern vector $\bm{u}$, for each element $\bm{v_i}$ we can compute the attention score $\alpha_i=a(\bm{u},\bm{v_i})$. This can also be termed as external attention, since attention is computed by matching an external pattern $\bm{u}$ with each element $\bm{v_i}$, and each score $e_i$ indicates quality of match. On the contrary, in self-attention, the external pattern $\bm{u}$ is replaced by parts of the sequence itself, and therefore is also termed as internal attention. To illustrate this with an example: \textit{Volleyball match is in
progress between ladies}, here \textit{match} is the sentence head on which all other tokens depend, and ideally we want to use self-attention to capture such intrinsic dependency automatically. Alternatively, we can interpret self-attention as matching with each element $\bm{v_i}$ an internal pattern $\bm{v'}$ within $V$:
\begin{equation}
e_i=a(\bm{v'},\bm{v_i})
\end{equation}
A typical choice for $\bm{v'}$ is simply another element $\bm{v_j}$, so as to compute a pairwise attention score, but in order to fully capture complex interaction between terms within a sequence, we can further extend this to compute attention between every pair of terms within a sequence, i.e., to set 
$\bm{v'}$ as each element $\bm{v_j}$ in a sequence and compute a score for each pair of terms. Therefore we modify the previous equations to:
\begin{eqnarray}
e_{ij}=a(\bm{v_i},\bm{v_j})\\
\alpha_{ij} = softmax(e_{ij})
\end{eqnarray}aiming to capture complex interaction and dependency between terms within a sequence. Then the choice of the alignment function $a$ is literally the same as the basic attention, such as a single layer neural network:
\begin{equation}
\alpha_{ij} = 
softmax(tanh (\bm{w^T}[\bm{v_i};\bm{v_j}] + b))
\end{equation}
In this way, each token maintains a distributed relation representation with respect to all other tokens and the complex pairwise relationship can be easily interpreted from its assigned scores. And the model can further be enriched with multi-dimensional attention \cite{shen2017disan} as we mentioned earlier.\\
Another motivation for self-attention is related to the word embedding. To be specific, we want to utilize self-attention models to learn complex contextual token representation in a self-adaptive manner. We can illustrate this point by an example of word sense disambiguation:\\ 
\textit{I arrived at the bank after crossing the street.}\\ 
\textit{I arrived at the bank after crossing the river.}\\
The word \textit{bank} has different meanings under different contexts, and we want our model to learn contextual token embeddings that can capture semantic information from their surrounding contexts. Transformer \cite{vaswani2017attention} is an exemplar novel attention-based architecture for machine translation. It is a hybrid neural network with sequential blocks of feed forward layers and self-attention layers. Similar to the previous self-attention mode, the novel self-attentive encoding can be expressed as:
\begin{eqnarray}
A = softmax[\frac{(VW_1)(VW_2)^T}{\sqrt{d_{out}}}]\\
C = A^T(VW_3)
\end{eqnarray}
Here $V = \{\bm{v_i}\} \in \mathbb{R}^{n \times d_{in}}$ represents an input sequence and $W_1, W_2,W_3 \in \mathbb{R}^{d_{in} \times d_{out}}$ are matrices to be learned for transforming $V$ to its query, key and value representation. $C = \{\bm{c_i}\} \in \mathbb{R}^{n \times d_{out}}$ therefore forms a sequence of self-attentive token encoding. We can expect each input token to learn a deep context-aware configuration via adjusting its relation with its surroundings during end-to-end training. We should also be aware that the architecture excludes all recurrent and convolution layers, as computation within a self-attention layer is parallel (therefore outweighs RNN) and parameter-efficient (compared with CNN). Various techniques have been proposed to further enhance its representation power. Positional encoding \cite{vaswani2017attention} is introduced to provide the model with additional positional information of each token, an example of which can be constructed as follows: 
\begin{eqnarray}
PE(pos, 2i) = sin (pos/10000^{2i/d})\\
PE(pos, 2i+1) = cos (pos/10000^{2i/d})
\end{eqnarray}
and later incorporated into the sentence encoding as additional features. To avoid receving attention from undesired direction, a directional mask $M$ \cite{shen2017disan}, a triangular matrix with -$\infty$ entries on the disabled position and $0$s otherwise, is added to the score representation before normalizing:
\begin{eqnarray}
\alpha_{ij} = softmax(e_{ij} + M_{ij})
\end{eqnarray} where 
\begin{equation}
M_{ij} = \begin{cases}
0, &i>j\\
-\infty, & \text{otherwise}
\end{cases}
\end{equation} for backward disabling. This can be useful while training a left-to-right language model since the future context should not provide any clue for generating the next token. Other techniques include relative position encoding \cite{shaw2018self} that aims for incorporating pairwise distance information into the contextual token representation $\bm{c_{i}}$ as following:
\begin{equation}
\bm{c_i} =\sum_{j}{\alpha_{ij}(\bm{v_j} + \bm{w_{ij}})}
\end{equation} where each weight $\bm{w_{ij}}$ corresponds to a directed edge from vertex $\bm{v_i}$ to $\bm{v_j}$, if $V$ is considered as a fully connected graph. These weights are initialized and learned during training.\\
Besides the pair-wise score computation, Lin et al. proposed an alternative method to obtain self-attention scores based on a fully connected neural network:
\begin{equation}
A = softmax(W_2tanh(W_1(V^T))
\end{equation} where each row in $A$ represents a single type of attention. Then the entire sequence encoding $C$ can be obtained by:
\begin{equation}
C = AV
\end{equation} In this case, the attention that an element receives is determined by its relevance to all elements in the entire sequence via full connection. Such a technique can be used for obtaining a fix-length encoding from a variable-length sequence since the dimension of $C$ is independent of the input sequence length.
  
\subsection{Memory-based Attention}
To introduce a new type of attention, we first reconstruct the old attention in an alternative way. Given a list of key value pairs $\{(\bm{k_i},\bm{v_i})\}$ stored in memory and a query vector $\bm{q}$, we redefine the three steps as: 
\begin{enumerate}
\item $e_i=a(\bm{q},\bm{k_i})$ (address memory)
\item $\alpha_i=\frac{exp⁡(e_i)}{\sum_{i}{exp(e_i)}}$  (normalize)
\item $\bm{c}=\sum_{i}{\alpha_i \bm{v_i}}$ (read contents)
\end{enumerate} Here we re-interpret computing attention score as soft memory addressing using query $\bm{q}$, and encoding as reading contents from memory based on attention scores $\{\alpha_i\}$, which constitutes the very basic form of memory-based attention. In fact, in quite a few literatures, "memory" is simply a synonym for the input sequence. Also note that if every pair of $\bm{k_i}$ and $\bm{v_i}$ are equal, this reduces to the basic attention mechanism. However, the alternative (memory-based) attention mechanism can become much more powerful as we incorporate additional functionalities to enable reusability and increase flexibility, both of which we will later discuss in detail. 
\subsubsection{Reusability}
A fundamental difficulty in some question answering tasks is that the answer is indirectly related to the question and therefore can not be easily solved via basic attention techniques (demonstrated at the beginning of this section). However, this can be achieved if we can simulate a temporal reasoning procedure by making iterative memory updates (also called multi-hop) to navigate attention to the correct location of the answer step-by-step \cite{sukhbaatar2015end}, an outline of which is illustrated in Figure \ref{memory-based attention}. Intuitively, in each iteration, the query is updated with new contents, and updated query is used for retrieving relevant contents. A pseudo run on an early example is given in Figure \ref{q_a}, 
where the query is initialized as the original question and is later updated by simply summing up the current query and content \cite{sukhbaatar2015end}:
\begin{equation}
\bm{q^{(t+1)}} = \bm{q^{(t)}} + \bm{c^{(t)}}
\end{equation}
More sophisticated update methods include constructing a recurrent network across query and content of multiple time steps \cite{kumar2016ask}, or inducing the output based on both content and location information \cite{graves2014neural}. 
Results show that when complex temporal reasoning tasks are given (similar to Figure \ref{q_a}), the memory-based attention model can successfully locate the answer after several hops.
\subsubsection{Flexibility}
\begin{figure}
\includegraphics*[width = 0.4\textwidth]{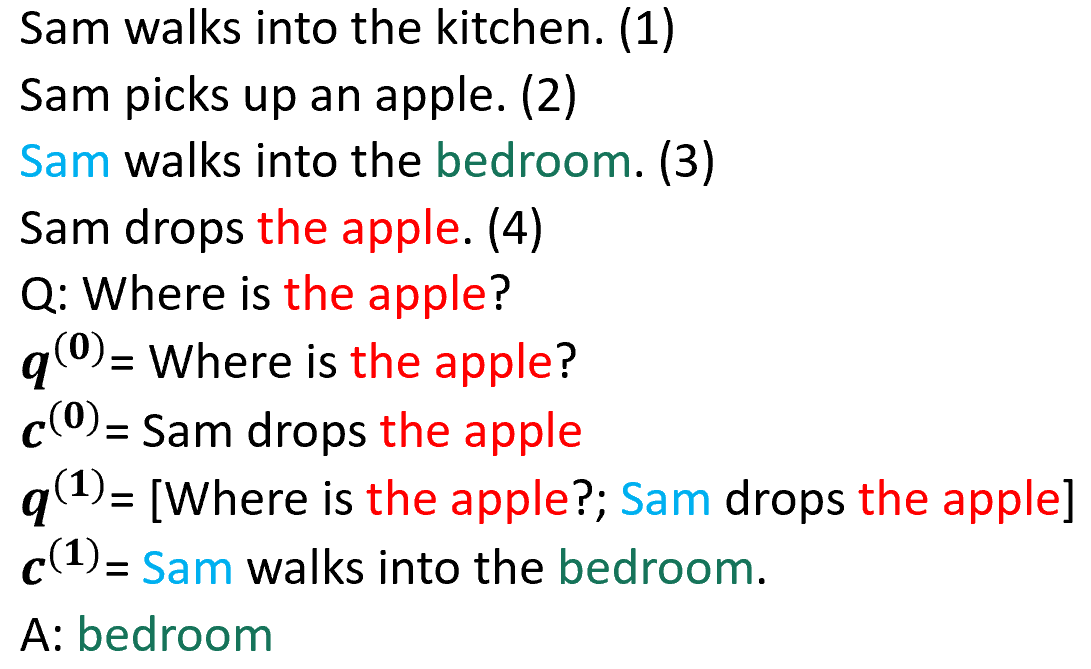}
\caption{A pseudo illustration of memory-based attention updates on an early question answering example.}
\label{q_a}
\end{figure}

Since keys and values are distinctly represented, we have freedom to incorporate prior knowledge in designing separate key and value embeddings to allow them to better capture relevant information respectively. To be specific, key embeddings can be manually designed to match the question and value embeddings to match the response. In key-value memory network \cite{miller2016key}, a window level representation is proposed such that keys are constructed as windows centered around entity tokens and values are those corresponding entities, aiming for more efficient and accurate matching. Then for the example in Figure \ref{q_a}, entities such as \textit{apple} and \textit{bedroom} are value embeddings, and their surrounding tokens are key embeddings.\\ 
More sophisticated architectures include the Dynamic Memory Network \cite{kumar2016ask} where the overall architecture is fine-split into four parts: question module, input module, episodic memory module and answer module, each of which is a complex neural micro-architecture itself. Such modularized design enables piecewise domain knowledge injection, efficient communication among modules, and generalization to a wider range of tasks beyond traditional question answering. A similar architecture is proposed to handle both textual and visual question answering tasks \cite{xiong2016dynamic}, where visual inputs are fed into a deep convolutional network and high-level features are extracted and processed into an input sequence for the attention network. If we further extend memory and query representation to fields beyond question answering, memory-based attention techniques are also used in aspect and opinion term mining \cite{wang2017coupled} where query is represented as aspect prototypes, in recommender systems \cite{zheng2018mars} where users become the memory component and items become queries, in topic modelings \cite{zeng2018topic} where latent topic representation extracted from a deep network constitutes the memory, etc.
\begin{figure}
\centering
\begin{enumerate}
 \item $initialize$  $\bm{q} = question$
 \item $e_i=a(\bm{q},\phi_k(\bm{k_i}))$ (address the memory)
 \item $\alpha_i = \frac{exp⁡(e_i)}{\sum_{i}{exp(e_i)}}$ (normalize)
 \item $\bm{c}=\sum_{i}{\alpha_i\phi_v(\bm{v_i})}$ (retrieve contents)
 \item $\bm{q} = update \_query(\bm{q},\bm{c})$ (update query)
 \item $goto$  2 (multi-hop)
\end{enumerate}
\caption{An enhanced version of memory-based attention with multi-hop and alternative key value embeddings}
\label{memory-based attention}
\end{figure}
\subsection{Task-specific Attention}
In this section we include alternative usage of attention that are intricately designed for a specific task. Although not as generalizable as methods introduced earlier, they are well-motivated and fit properly into their own task, therefore worth mentioning. Tan et al. proposed an alternative for computing attention scores for the task of abstractive document summarization \cite{tan2017abstractive}. The formulation of new graph-based attention is similar to PageRank algorithm \cite{page1999pagerank}. Given a document $V = \{\bm{v_i}\}$ where each $\bm{v_i}$ represents a sentence, the stationary attention distribution $\bm{\alpha} = \{\alpha_i\}$  satisfies:
\begin{equation}
\bm{\alpha} = \lambda WD^{-1}\bm{\alpha} + (1-\lambda)\bm{y}
\end{equation}
Here $W$ is a square matrix in which each entry $W_{ij}$ encodes a multiplicative composition between $\bm{v_i}$ and $\bm{v_j}$, and $D$ is a diagonal normalizing matrix to ensure each column of $WD^{-1}$ sums up to 1. $\lambda$ and $\bm{y}$ are the damping factor and an auxiliary uniform distribution, respectively. Then $\bm{\alpha}$ can be solved analytically as:
\begin{equation}
\bm{\alpha} = (1- \lambda)(I - \lambda WD^{-1})^{-1}\bm{y}
\end{equation}
The underlying motivation is that a sentence is important in a document if it is heavily linked with many important sentences.\\
Kim et al. proposed a structured attention network which integrates attention mechanism with probabilistic graphical model by introducing a sequence of discrete latent variables $\mathbf{Z} = \{\mathbf{z_i}\}$ as soft selectors into the input sequence \cite{kim2017structured}. An attention distribution $p(\mathbf{Z}=\bm{z}|V,\bm{q})$ is generated from a conditional random field and then used to encode the context vector as an expectation over this probability distribution:
\begin{equation}
\bm{c}=E_{\bm{z} \sim p(\bm{z}|V,\bm{q})}[f(V,\bm{z})]
\end{equation} where $f(V,\bm{z})$ is the annotated function that models relationship between latent variables and the given inputs. If $\mathbf{Z}$ is a single random variable and given $f(V,\mathbf{Z} = z) = V_z$ (i.e., selecting the $z^{th}$ element from $V$), then this is equivalent to soft selection as in the basic attention. Without this restriction, they demonstrate its adaptability to multiple input selection tasks such as syntactic tree selection and subsequence selection under a general case.\\
In machine translation, a local attention model \cite{luong2015effective} is proposed to handle long sequence translation where computation of global attention (i.e. attending to every element) is expensive. During decoding, a pivot position $p_t \in [0, length(V)]$ , which specifies the center of attention, is first predicted by a small neural network, and then Gaussian Smoothing is applied around the center to produce soft alignment. 

\section{Application}
In the previous section, we have showed that attention along with its variants have been widely applied to various NLP tasks. Here we will further explore the connection of attention to other abstract concepts in machine learning. As we have discussed previously, attention can be used for \textit{encoding} a sequence by extracting important terms based on its match with a given pattern; attention can also be used for iterative memory \textit{addressing} and reading given a query. Here we present  three more applications of attention: \textit{emsemble}, \textit{gating}, and \textit{pre-training}. 
\subsection{Attention for Ensemble}
If we interpret each element $\bm{v_i}$ in a sequence as an individual model, and normalized scores $\alpha_i$ as their weighted votes, applying the attention mechanism can then be analogous to model ensemble. This is explored in Kieta et al. where they ensemble a set of word embeddings to construct a meta-embedding with more representative power and flexibility \cite{kiela2018dynamic}. Specifically, attention score $\alpha_{i}^{[j]}$ for the embedding $\bm{v_{i}^{[j]}}$ (the $i^{th}$ embedding for the $j^{th}$ word) is given via self-attention:
\begin{equation}
\alpha_{i}^{[j]} = softmax(\bm{w^{T}v_{i}^{[j]}}+b)
\end{equation}
And the meta-embedding $\bm{v^{[j]}}$ for the $j^{th}$ word is given by: 
\begin{equation}
\bm{v^{[j]}} = \sum_{i}{\alpha_{i}^{[j]}\bm{v_{i}^{[j]}}}
\end{equation} They demonstrate that certain embeddings are preferred over others depending on characteristics of the word, such as concreteness and frequency. For example, the ImageNet embedding \cite{he2016deep} receives larger weights than FastText embeddings \cite{bojanowski2016enriching} for concrete words.  
\subsection{Attention for Gating}
Another application of attention is to integrate this mechanisms with memory updates in recurrent network. In traditional GRU \cite{chung2014empirical}, hidden state updates are given by:
\begin{eqnarray}
\tilde{\bm{h_i}} = tanh(W\bm{v_i} + \bm{r_i} \circ (U\bm{h_{i-1}}) + \bm{b^{(h)}})\\
\bm{h_i} = \bm{u_i} \circ \bm{\tilde{h_i}} + (\bm{1} - \bm{u_i}) \circ \bm{h_{i-1}}
\end{eqnarray}
where $\bm{u_i}$ and $\bm{r_i}$ are update and reset gates learned during training. While in an alternative attention-based GRU \cite{xiong2016dynamic}, $\bm{u_i}$ is replaced by a scalar attention score $\alpha_i$ received by the $i^{th}$ elelment when updating its hidden state. Then the last update step can be replaced by:
\begin{equation}
\bm{h_i}=\alpha_i \circ \tilde{\bm{h_i}} + (1 - \alpha_i) \circ \bm{h_{i-1}}
\end{equation} The attention scores are computed in an external module. Such an attention-based gating allows context-aware updates based on global knowledge of previous memory, and easier interpretability of importance of each element.\\Similarly in text comprehension, memory-based attention gate $\bm{\tilde{q_i}}$ is constructed \cite{dhingra2016gated}  based on the interaction between the query $Q$ and each token $\bm{v_{i}}$ in the document and iteratively update each token embeddings:
\begin{eqnarray}
\alpha_i = softmax(Q^T\bm{v_{i}})\\
\bm{\tilde{q_i}}=Q\alpha_i\\
\bm{\tilde{v_{i}}} = \bm{v_{i}} \circ \bm{\tilde{q_i}}\\
\bm{v_i},Q = GRU_v(\bm{\tilde{v_{i}}}),GRU_Q(Q)
\end{eqnarray}
aiming to build up deep query-specific token representation. 
\subsection{Attention for Pre-training}
Pre-trained word embeddings are crucial to many NLP tasks. Traditional methods such as \textit{Skipgram, Cbow, and Glove} \cite{mikolov2013efficient,pennington2014glove,mikolov2013distributed} take use of large text corpora to train an upsupervised prediction model based on contexts and learn a high dimensional distributed representation of each token. On the contrary, recently proposed pre-training methods integrate attention-based techniques with deep neural architectures, aiming to learn higher quality token representation that incorporates syntactic and semantic information from the surrounding contexts. and then the model is fine-tuned to adapt to a downstream supervised task. BERT \cite{devlin2018bert} is a bi-directional pre-training model backboned by the Transformer Encoder \cite{vaswani2017attention}, a deep hybrid neural network with feed forward layers and self-attention layers which we have briefly discussed in section \ref{self-attention}. During pre-training, one task is to learn a bi-directional masked language model, meaning that a small percent of tokens in a sentence are masked and the goal is to predict these tokens based on their context. The other task is binary next sentence prediction, where two spans of texts are sampled from the corpora and the model is trained to predict whether they are contiguous. As discussed in Section \ref{self-attention}, each token redistributes attention across the sequence and reconstruct its interaction with other tokens in a self-adaptive manner as the training proceeds, aiming to learn its contextual representation based on the entire sequence. When the pre-trained model is integrated with a supervised task, an additional layer is added on top of the model and fine-tuned to adapt to its supervised downstream task. The new model has achieved ground-breaking results on various NLP tasks, by focusing on pre-training the deep hybrid architecture on large text corpora and then sparing minimal efforts on fine-tuning. Other attention-based pre-training models include OpenAI GPT, which instead uses a Transformer Decoder (with backward disabling mask) to pre-train a deep left-to-right language model based on a different set of tasks. 
\section{Evaluation}
Compared to the universal usage of attention, only a few attempts are made either to give a rigorous mathematical justification of why it works in various scenarios. Nevertheless, there are several works that have attempted to set up standards on evaluating its performance, either qualitatively or quantitatively, task-specific or general, and here we give a short summarization of these approaches.
\subsection{Quantitative}
\begin{figure}
\centering
\includegraphics*[width = 0.4\textwidth]{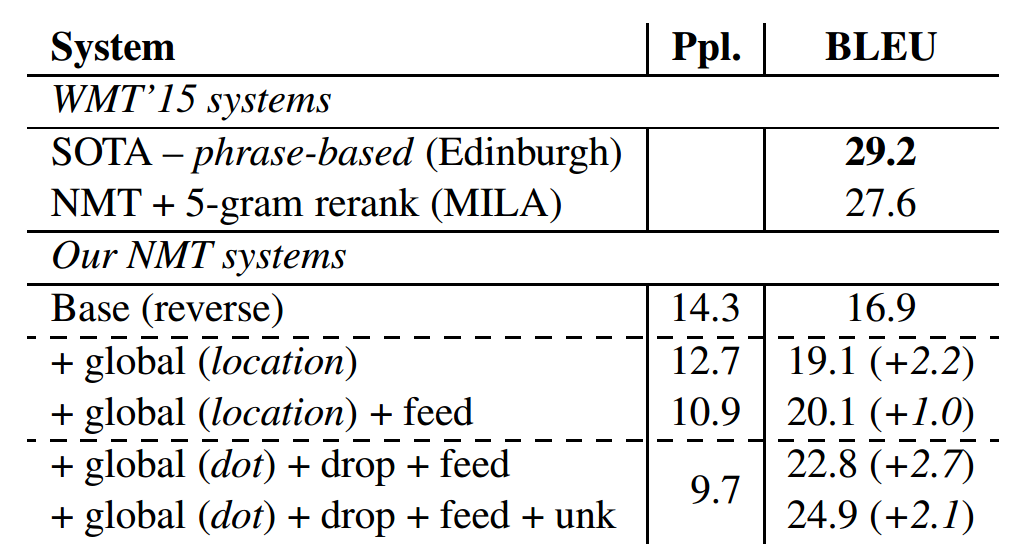}
\caption{An illustration of extrinsic evaluation of different attention mechanisms under a downstream machine translation task. The image is modified from \cite{luong2015effective}}.
\label{quantitative}
\end{figure}

Quantitative evaluation on attention can be further divided into intrinsic or extrinsic based on whether the contribution of attention is assessed on itself or along within a downstream supervised task.\\Intrinsic evaluation methods are typically proposed in machine translation \cite{luong2015effective}, where attention is analogous to word alignment, performance of attention could be directly measured by comparing the attention distribution with the gold alignment data, and quantified using alignment error rate (AER). Similarly, Liu et al. proposed a method to manually construct "gold attention vectors" \cite{liu2017exploiting} by first identifying labelled key words within a sentence and then conducting post-processing procedures such as smoothing and normalization, given abundant well-annotated data. For example, for the sentence \textit{\textbf{Mohamad} fired \textbf{Anwar}, his former \textbf{protege}, in \textbf{1998}}, the four tokens in boldface are labelled as argument words and receive an attention score of $0.25$ each ($0.25 \times 4 = 1$), and then smoothing is optionally applied around each attended token.  Though intrinsic evaluation methods produce precise measurements on performance, they tend to be restricted to their specific tasks and rely heavily on plentitude of labelled data. \\On the other hand, extrinsic evaluation methods (Figure \ref{quantitative}) are more general and widely used. This can be easily formulated by comparing the overall performance across different models under the downstream task. However, the result could be misleading since whether or not the improvements should be attributed to the attention component can not be determined.

\subsection{Qualitative}
\begin{figure}
\centering
\includegraphics*[width = 0.4\textwidth]{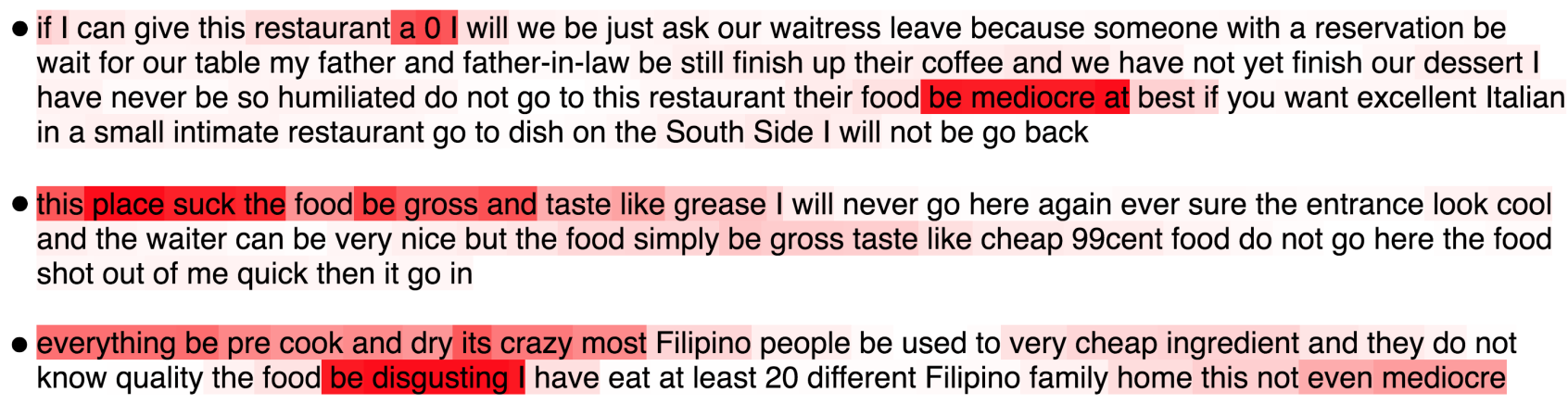}
\caption{An illustration of qualitative methods based on heatmaps. Higher color intensity indicates higher attention score. The image is modified from \cite{lin2017structured}}.
\label{qualitative}
\end{figure}

Qualitative evaluation for attention is currently the most widely used evaluation technique, due to its simplicity and convenience for visualization (Figure \ref{qualitative}). To be specific, a heat-map is constructed across the entire sentence where the intensity is proportional to the normalized attention score each element receives. Intuitively, attention is expected to be focused on key words for the corresponding task. However, such an approach turns out to be better for visualization than for analysis.

\section{Conclusion and Prospects}
In this paper, we have surveyed through recent works on the attention mechanism and conducted an introductory summary based on its formulation, variation, application and evaluation. Compared to its wide usage in various NLP tasks, attempts to explore its mathematical justification still remain scarce. Recent works that explore its application in embedding pre-training have attained great success and might be a prospective area of future research.

\bibliographystyle{aaai}

\end{document}